\pdfoutput=1 

\documentclass[letterpaper]{article}
\usepackage{tex_macros}
\usepackage{makeidx}
\usepackage{etoolbox}
\usepackage{array}
\usepackage{siunitx}
\usepackage[itemlayout = relhang]{idxlayout}
\usepackage{changes}
\usepackage{titling}
\let\mathcal\relax
\usepackage[cal=cm]{mathalfa}
\usepackage{algorithm}
\usepackage[]{algpseudocode}
\usepackage{pifont}
\usepackage{mathtools}
\mathtoolsset{showonlyrefs} 
\usepackage{graphicx}

\algnewcommand{\Inputs}[1]{%
  \State \textbf{Inputs:}
  \Statex \hspace*{\algorithmicindent}\parbox[t]{.8\linewidth}{\raggedright #1}
}
\algnewcommand{\Initialize}[1]{%
  \State \textbf{Initialize:}
  \Statex \hspace*{\algorithmicindent}\parbox[t]{.8\linewidth}{\raggedright #1}
}

\algnewcommand{\Given}[1]{%
  \State \textbf{Given:}
  \Statex \hspace*{\algorithmicindent}\parbox[t]{.8\linewidth}{\raggedright #1}
}

\let\oldtextbf=\textbf
\renewcommand{\textbf}[1]{\ifmmode\mathbf{#1}\else\oldtextbf{#1}\fi}

\renewcommand{\textit}[1]{\emph{#1}}

\newcommand{\xmark}{\ding{53}}%

\newcolumntype{L}[1]{>{\raggedright\let\newline\\\arraybackslash\hspace{0pt}}m{#1}}
\newcolumntype{C}[1]{>{\centering\let\newline\\\arraybackslash\hspace{0pt}}m{#1}}
\newcolumntype{R}[1]{>{\raggedleft\let\newline\\\arraybackslash\hspace{0pt}}m{#1}}

\makeatletter
\renewcommand*\env@matrix[1][*\c@MaxMatrixCols c]{%
  \hskip -\arraycolsep
  \let\@ifnextchar\new@ifnextchar
  \array{#1}}
\makeatother
\usepackage{uai2019}
\usepackage[margin=1in]{geometry}

\usepackage{times}

\title{Learning Gaussian Policies from Corrective Human Feedback}

\author{} 

\author{ {\bf Daan~Wout\thanks{~D. Wout, J. Scholten, C. Celemin and J. Kober are with the Department of Cognitive Robotics at the Delft University of Technology, the Netherlands. \textproc{daan090@gmail.com, jan@jjscholten.nl, \{c.celemin,j.kober\}@tudelft.nl}}} \\
\And
{\bf Jan~Scholten}  \\
\\
\And
{\bf Carlos~Celemin}   \\
\\
\And
{\bf Jens~Kober}   \\
\\
}

\begin{document}

\maketitle

\begin{abstract}
\acresetall
Learning from human feedback is a viable alternative to 
control design that does not require modelling or control expertise.
Particularly, learning from corrective advice garners advantages
over evaluative feedback as it is a more 
intuitive and scalable format.
The current state-of-the-art in this field, \acs{COACH}, 
has proven to be a effective approach 
for confined problems.
However, it parameterizes the policy with Radial Basis Function networks, which require meticulous feature space engineering for higher order systems.
We introduce \acf{GPC},
where feature space engineering is avoided by employing \aclp{GP}.
In addition, we use the available policy uncertainty 
to 1) inquire feedback samples of maximal utility and 2) to adapt the learning rate to the teacher's learning phase.
We demonstrate that the novel algorithm outperforms the current state-of-the-art in final performance, convergence rate and robustness to erroneous feedback in OpenAI Gym continuous control benchmarks, both for simulated and real human teachers.
\acresetall
\end{abstract}

\section{INTRODUCTION}
            \begin{figure}
                \centering
                \includegraphics[trim=1 25 9 0, clip,width = \linewidth]{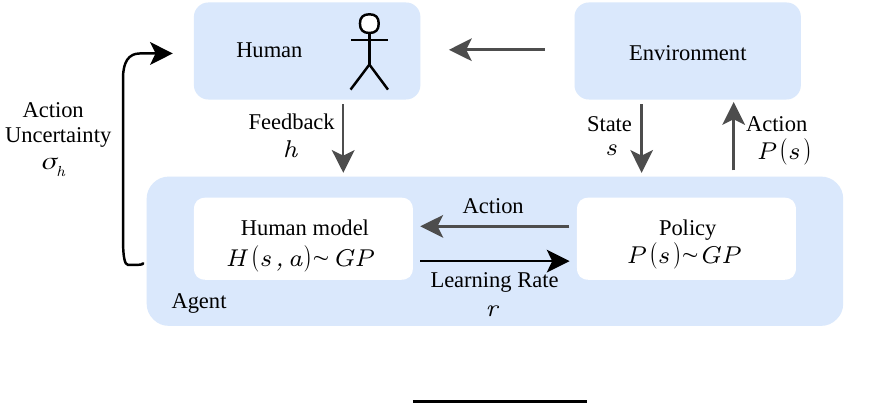}
                \caption{A schematic diagram of the feedback framework of \acf{GPC}. The teacher provides feedback $h$ to corrected the observed action in the respective state.}
                \label{fig:beginpage}
            \end{figure}
            
            In contrast to autonomous Machine Learning techniques, humans 
            are very effective in inferring suitable control strategies when facing new problems. Specifically for intuitive problems, like picking up objects or playing simple games, humans are able to achieve decent performance on first try \citep{hessel2018rainbow}. 
            Communicating this domain knowledge has shown to drastically accelerate model-free control techniques.
            For example, %
            a well known approach is the \ac{LfD} framework, where a policy is derived using examples of proper execution \citep{ross2011reduction}. Other methods, like Apprenticeship Learning, employ demonstration to reversely derive the trainer's choices for autonomous improvement \citep{abbeel2004apprenticeship}.
            To avoid superfluous (and possibly expensive) interaction with the trainer, \cite{grave2013learning, losey2018including} improved sample efficiency 
            with an \ac{AL} framework where
            demonstrations are inquired especially for uncertain policy executions.

            \ac{LfD} could however be troublesome for systems that feature agile dynamics. 
            Moreover, the demonstrations require expert knowledge of the system and the solution \citep{argall2009survey}. 
            A less demanding 
            approach has been studied by, e.g., \cite{griffith2013policy,knox2009interactively}, where the trainer gives scalar reward signals (\emph{evaluative} feedback) in response to the agent's observed behavior.
            \cite{thomaz2006reinforcement} however argue that trainers implicitly guide in their reward signal, and base their feedback not solely on past actions but also on what is going to happen.
            This intrinsic preference in guidance has been studied by \cite{celemin2015coach} and resulted in \ac{COACH}, an algorithm that allows teachers to shape the optimal policy by providing \emph{corrective} feedback, i.e.\ in the action domain. 
            This approach engages the users intuition without requiring expertise on the task. 
            Moreover, teachers are now able to guide a policy rather than to evaluate it, which has shown to be better scalable to high-dimensional problems \citep{suay2011effect}.
            \ac{COACH} has shown to be very efficient on 
            intuitive
            problems and outperforms evaluative approaches in human assisted learning.
            However, \ac{COACH} employs \ac{RBF} networks, which require extensive design procedures. The application of \ac{COACH} is therefore limited  to simple and confined problems and hence does not exploit the full potential of corrective feedback implementations.

            In order to improve on this point,
            we introduce \ac{GPC}, a corrective feedback framework following \ac{COACH}'s structure but comprising engineering advantages by introducing \acp{GP} as alternative to \ac{RBF} networks.
            In addition, we employ the available uncertainty estimations of \acp{GP} for 1) A pioneering approach of employing \ac{AL} in an corrective feedback framework, analogue to what \cite{maeda2017active,grave2013learning} do with \ac{LfD}. 2) Match the learning rate to the learning phase of the teacher and adapt policy corrections accordingly.
            We apply the novel algorithm to several benchmarks of the OpenAI gym \citep{openaigym}, both with simulated and real human teachers, to show the significance of the proposed contributions and the performance in a comparison against previous work.
            
            This study is organized as follows: background material is covered in \sref{sec:background}. \sref{sec:GPC} details the novel algorithm \ac{GPC}. The experimental setup and corresponding results are presented in \sref{sec:expsetup} and \sref{sec:results} respectively.

\section{BACKGROUND}\label{sec:background}
In the following, we detail the key components of \acs*{GPC}, starting with \acs*{COACH} which is the basis of the novel framework. The principles of \ac{GP} are covered thereafter.

\subsection{COACH}\label{sec:coach}

\begin{algorithm}
  {
  \caption{~\ac{COACH} framework}
  \label{tab:COACH}
  \begin{algorithmic}[1]
    \Given{\strut Policy learning rate $e$ \\
    Human model learning rate $\beta$\\
    Constant learning rate $c_{c}$\\
    Feature space function $\phi(\cdot)$}
    \For{\text{all} $k$ }
      \State Get state $\mathbf{s_k}$
      \State Compute new action $a_k \gets \theta_k \phi(\mathbf{s_k})$
      \State Obtain corrective advise $h$ 
      \If{$h \neq 0$}
            \State $ H(\mathbf{s_k}) = \psi_k^T \phi(\mathbf{s_k})$ 
            \State $\Delta \psi = \beta (h-H(\textbf{s}_k)) \phi(\textbf{s}_k)$ 
            \State Human model update $\psi_{k+1} = \psi_k +  \Delta \psi $
            \State Get learning rate $\alpha(\textbf{s}_k) = |H(\textbf{s}_k)| + c_c$
            \State $\Delta \theta  = \alpha(\textbf{s}_k) \phi(\textbf{s}_k) h e$
            \State Policy update $\theta_{k+1} = \theta_k +  \Delta \theta$
      \EndIf
    \EndFor
  \end{algorithmic}
  }
\end{algorithm}

\acf{COACH}, proposed by \cite{celemin2015coach}, is an algorithm that trains agents with corrective advice. It has policy $P_c:S\rightarrow \mathbb{R}^n$, with $S$ the set of states and $n$ the action-space dimensionality, that maps states to continuous actions. The trainer observes the agent and occasionally suggests to either increase or decrease the action. 
This feedback $h\in \{-1,1\}$ is modelled in the \textit{human feedback model}: $H_c:S\rightarrow \mathbb{R}^n$.
The parameterization of both models is done by \ac{RBF} networks, where the respective models have different weight to the feature vector $\phi(\textbf{s}_k)$, with $\textbf{s}_k$ denoting the state in time-step $k$. 
The learning framework is supported by the following modules.

\subsubsection{Policy Supervised Learner}\label{sec:policysupervisedlearner}

The policy $P_c(\textbf{s}_k)$ provides the action $a$ for a given state $\textbf{s}_k$, by taking the linear combination of the weights and the feature vector, i.e.\ $a_k = P_c(\textbf{s}_k) = \theta_k^T \phi(\textbf{s}_k)$,
with $\theta$ the weight vector of the policy. For every directive correction $h$ given by the teacher, the weight vector is updated according to a Stochastic Gradient Descent approach:
\begin{equation}
    \theta_{k+1} = \theta_k - \alpha(\textbf{s}_k) \nabla_\theta J(\theta),
\end{equation}
with $\alpha(\textbf{s}_k)$ the learning rate (obtained as described in \sref{sec:coachhuman}) and $J(\theta)$ denoting the cost function, which is the squared error between the applied and 'desired' action, given by $h$ and magnitude $e$. The latter denotes a free parameter set by the user within the range of the action domain.
Hence, taking the human feedback into account, the gradient becomes
\begin{equation}\label{eq:updateparampolcoach}
    \theta_{k+1} = \theta_{k} + \alpha(\textbf{s}_k) \phi(\textbf{s}_k) h e.
\end{equation}

\subsubsection{Human Feedback Supervised Learner}\label{sec:coachhuman}

This module models the feedback of the trainer as a function of the state $\textbf{s}_k$. 
The predictions are given by a linear combination of the human model weights $\textbf{\psi}_k$ and the feature vector $\phi(\textbf{s}_k)$, i.e.\ $H(\textbf{s}_k) = \textbf{\psi}_k \phi(\textbf{s}_k)$.
The updates on the weight vector $\textbf{\psi}_k$ are conducted in the same fashion as \eref{eq:updateparampolcoach}, but now with a known error magnitude $e_h = h - H(\textbf{s}_k)$ such that
\begin{equation}
  \textbf{\psi}_{k+1} = \textbf{\psi}_{k} + \beta  (h-H(\textbf{s}_k)) \phi(\textbf{s}_k)
\end{equation}
with $\beta$ the learning rate of the human model. 
By this human model, the learning rate in \eref{eq:updateparampolcoach} is given as
\begin{equation}\label{eq:learningratec}
    \alpha(\textbf{s}_k) = |H(\textbf{s}_k)| + c_c.
\end{equation}
Note that $H(\textbf{s}_k) \approx 1$ for consistent feedback with equal sign and hence increases the learning steps. For alternating feedback the learning rate diminishes. To prevent the learning rate from dwindling to zero, \eref{eq:learningratec} is appended with a constant factor $c_c$.

The outline of the \ac{COACH} framework is depicted in \aref{tab:COACH}. Lines~3-5 comprise of policy executions. The update lines consist of the human prediction and human model updates (line 7-10) as motivated in \sref{sec:coachhuman}. The policy updates from \sref{sec:policysupervisedlearner} are subsequently given in lines~11-12. Furthermore, the \ac{COACH} framework can be extended by the \textit{Credit Assigner}, which takes a human delay into account for the feedback given by the teacher. Since this study will not exploit this feature it will not be covered here.

\subsection{GAUSSIAN PROCESSES}    
    \acfp{GP} are Bayesian non-parametric function approximation models. It is a collection of random variables, such that every finite collection of those random variables has a multivariate normal distribution. 
    \acp{GP} do not require specification of a model structure a priori and provide the uncertainty along with the predictions. 
    A \ac{GP} is fully specified by its mean $m(\textbf{x})$ and covariance function $k(\textbf{x},\textbf{x'})$, i.e.
    \begin{equation}
        f(\textbf{x}) \sim \mathcal{GP}(m(\textbf{x}), k(\textbf{x},\textbf{x'})).
    \end{equation}
    Let $\textbf{y} = \{y_1,...y_n\}$ be a set of observations from a stochastic process
    \begin{equation}\label{eq:stochproc}
        y_i = f(\textbf{x}_i)+\epsilon,
    \end{equation}
    where $\textbf{x}_i$ denotes the input vector of observation $y_i$. 
    The noise $\epsilon$ is assumed Gaussian with standard deviation $\sigma_o$.
    The input matrix is defined as $X = \{\textbf{x}_1,...,\textbf{x}_n\}$.    
    Applying the conditional distributions \citep{rasmussen2006gaussian},
    the following posterior predictive equations for test inputs $\textbf{x}_*$ are given as:
    \begin{align*}
        \mathbf{f_*}|X,\mathbf{y},\textbf{x}_* & \sim \mathcal{N}(\mathbf{\bar{f}_*},\operatorname{cov}(\mathbf{f_*})), \quad \text{where} \\
        \mathbf{\bar{f}_*} &= m(\textbf{x}_*) + K_*[K+\sigma_n^2 I]^{-1}(\mathbf{y} - m(X)),\\
        \operatorname{cov}(\mathbf{f_*}) & = K_{**}-K_* [K+\sigma_n^2 I]^{-1} K_*, 
    \end{align*}
    where $K_* = k(X,\textbf{x}_*)$, $K_{**} = k(\textbf{x}_*,\textbf{x}_*)$, and $K$ is the Gram matrix with entries $K_{ij} = k(x_i,x_j)$. 
    The Gaussian noise per observation is denoted as $\sigma_n$ and has a similar function as $\epsilon$ in \eref{eq:stochproc}. 
    The kernel function $k(\textbf{x},\textbf{x}')$ is a measure of similarity between two input vectors $\textbf{x}$ and $\textbf{x}'$. 
    In this study, we employ two kernel functions. The first one is the squared exponential (SE) kernel, which is given as
    \begin{equation}\label{eq:sekernel}
        k_s(\textbf{x},\textbf{x}') = \sigma_s^2 \exp\left(-\frac{|\textbf{x}-\textbf{x}'|^2}{2l^2}\right),
    \end{equation}
    with $\textbf{\beta}_r = \{\sigma_s,l\}$ the \textit{hyperparameters} of the kernel function, consisting of the signal variance $\sigma_s$ and length-scale $l$. The length-scale denotes a measure for the roughness of the data. In general, one can assume that extrapolating more than $l$ units away from the input data is considered unreliable.
    The second kernel function, the Mat\'ern kernel, is specified as
    \begin{equation}\label{eq:matern}
        k_m(\textbf{x},\textbf{x}') = \sigma_m^2 \frac{2^{1-\nu}}{\Gamma(\nu)} \left( \sqrt{2\nu}\frac{2}{l}\right)^\nu B_v\left(\sqrt{2\nu}\frac{|\textbf{x}-\textbf{x}'|}{l} \right)
    \end{equation}
    with $B_v(\cdot)$ the modified Bessel function \citep{abramowitz1965handbook}, $\Gamma(\cdot)$ the Gamma function and the \textit{hyperparameters} $\textbf{\beta}_m = \{\sigma_m,l,\nu \}$. Here, $\nu$ denotes a `smoothness' parameter that correlates with the amount of times the target function is differentiable \citep{rasmussen2006gaussian}. 
    
    For multivariate targets, we train conditionally independent \acp{GP} for each target dimension. 

\section{GAUSSIAN PROCESS COACH (GPC)}\label{sec:GPC}
    We now introduce \acf{GPC}, 
    an algorithm based on \ac{COACH} that employs \ac{GP} as an alternative to \ac{RBF} networks to comprise advantage in scaling and sample efficiency.
    A schematic of the method is depicted in \fref{fig:beginpage}.
    In the main format, the trainer observes the environment and the current policy and provides action corrections to advance the policy. These corrections trigger agent updates in order to take immediate effect on the policy. This process is repeated until convergence.
    The pseudo-code of \ac{GPC} is in \aref{alg:GCOACH}.
    
    This section defines \ac{GPC} for a one dimensional \mbox{action-space}, but scales straightforwardly to higher dimensional problems. 

        \begin{algorithm}
        {
          \caption{\ac{GPC} Algorithm}
          \label{alg:GCOACH}
          \begin{algorithmic}[1]
            \Given{\strut Kernels $k(\cdot)$ for  $H$ and $P$ \\
            Hyperparameters $\beta$ with $M_{cs}$ or $M_{ns}$ \\
            Constant learning rate $c_r$}
            \vspace{.5em}
            \For{all $k$ within episode }
              \State Get state $\mathbf{s_k}$ 
              \State Execute action $a_k = P(\mathbf{s}_k)$ and obtain $\sigma_p(\mathbf{s}_k)$
              \State Obtain corrective advise $h \in \{-1,1\}$ 
              \If{$h \neq 0$}
                    \State $\mathbf{z_k} = (\textbf{s}_k,a_k)$
                    \State $h_\text{est} = H(\mathbf{z}_k)$ with uncertainty $\sigma_h(\mathbf{z}_k)$
                    \State Learning rate $r_k \gets \sigma_p(\textbf{s}_k) + \sigma_h(\textbf{z}_k)  + c_r$
                    \State New action $a_n = a_k + r_k \cdot h_k$ 
                    \State Update dictionary $P$ and apply \textproc{Spars}$(N_p)$
                    \State Update dictionary $H$: $N_h = \{...,(\textbf{z}_k,h_k)\}$
                    \State Update $M_{p},M_{h} \gets \operatorname{cov}(N_p,N_h)$ // NS only
                    \State Train \acp{GP}: \textproc{Train}$(P,H)$
              \EndIf
            \EndFor
          \end{algorithmic}
          }
        \end{algorithm}

    \subsection{MODELLING POLICY AND FEEDBACK}
    
        The \ac{GPC} framework engages two \ac{GP} models: the policy $P$ and the human model $H$.
        The prior of the policy is modelled as:
        \begin{equation}\label{eq:policymodel}
            P:S \rightarrow \mathbb{R} \sim \mathcal{GP}(m_p(\mathbf{s}),k_p(\mathbf{s},\mathbf{s'})),
        \end{equation}
        Here, $m_p(\mathbf{s})$ is assumed $0$.
        The policy is trained with the set $N_p =\{(\textbf{s}_1,a_1),(\textbf{s}_2,a_2),...,(\textbf{s}_m,a_m)\}$, which contains state-action data derived from the directional feedback from the trainer (details in \sref{sec:policyenhancement}).
        The human feedback is modelled by
            \begin{equation}\label{eq:humanmodel}
            H:S \times A \rightarrow \mathbb{R} \sim \mathcal{GP}(m_h(\mathbf{z}),k_h(\mathbf{z},\mathbf{z'})),
        \end{equation}
        with $A$ the action space. The mean is assumed $m_h(\mathbf{z}) = 0$.
        This human model is trained with the set
        $N_h = \{(\textbf{z}_1,h_1),(\textbf{z}_2,h_2),...,(\textbf{z}_v,h_v)\}$, where $\textbf{z}$ denote the concatenation of state $\textbf{s}$ and action $a$, and $h \in \{-1,1\}$ the suggested action correction of the teacher (decrease or increase). The proposed \ac{GPC} introduces a different human model with respect to the one of \ac{COACH}, where the feedback was only state dependent, i.e.\ $H_c:S\rightarrow \mathbb{R}$ (see \sref{sec:coach}). We have integrated the action in our human model to infer the human feedback per state-action, rather than state only. Further details on this principle are provided in \sref{sec:bayesianproperties}, where we elaborate on the uncertainty advantages. 
        
        Both models of \ac{GPC} require a kernel function that represents how the target function and uncertainty propagates along the input dimensions. 
        For the human model $H$ we assume a smooth propagation of the target function and therefore adopt the SE kernel \eref{eq:sekernel}. To allow for more freedom in the policy function in terms of roughness and discontinuities, we adopt the Mat\'ern kernel \eref{eq:matern} for $P$ \citep{rasmussen2006gaussian,duvenaud2014automatic}.

    \subsection{FEATURE SCALING}\label{sec:scaling}

        The policy in \eref{eq:policymodel} and human model in \eref{eq:humanmodel} both concern a multidimensional regression on the input data.
        Each input dimension may however be subject to data with completely different orders of magnitude, such that a single length-scale is unsuitable. 
        We therefore take an approach that allows us to set an independent length-scale per input dimension.
        
        Let us consider the SE kernel from \eref{eq:sekernel}. Following \cite{rasmussen2006gaussian}, the parameterization in terms of the \textit{hyperparameters} results in
        \begin{equation}
            k_s(\textbf{x},\textbf{x}') = \sigma_s^2 \exp\left(-\frac{1}{2}(\textbf{x}-\textbf{x}')^T l_sM (\textbf{x}-\textbf{x}')\right),
        \end{equation}
        with $M$ the diagonal matrix consisting of the \textit{characteristic length-scales} per axis. Such a covariance function implements \ac{ARD} \citep{neal1995bayesian}.    
        This study adopts two distinct methods for determining the diagonal values of $M$. In the first approach we let the trainer decide on the respective relevance of the input dimensions:
        \begin{equation}
            M_{cs} = \operatorname{diag}(\mathbf{w})^{-2},
        \end{equation}
        with $\textbf{w}$ a vector consisting of custom `weights' on the input dimensions. These values are determined a priori and deemed static throughout the learning process. This method is referred to as GPC(-CS).        
        The second method concerns the normalization of the independent inputs for an equal relative dependency, resulting in an approach where any length-scale tuning is circumvented. The result is an approach that does not scale with the input dimension and could hence be decisive for higher-order systems.
        The parameterization is carried out by
        \begin{equation}
            M_{ns} = \operatorname{diag}(\textbf{\sigma}_m)^{-2},
        \end{equation}
        with $\textbf{\sigma}_m$ the vector containing the variance of the independent input dimensions, which is updated for every feedback sample (see line 13 in \aref{alg:GCOACH}). 
        This method will be referred to as GPC-NS.
        
        The extension to the Mat\'ern kernel \eref{eq:matern} is straightforward with $M_{cs} = \operatorname{diag}(\mathbf{w})$ and $M_{ns} = \operatorname{diag}(\textbf{\sigma}_m)$ for every length-scale $l$. To distinguish between the scaling of the policy and the human model we add subscript $h$ and $p$, e.g.\ $M_{cs,h}$.
    \subsection{LEVERAGING UNCERTAINTY}\label{sec:bayesianproperties}
\begin{figure}
            \centering
            \includegraphics[trim=7 8 8 7, clip,width=\linewidth]{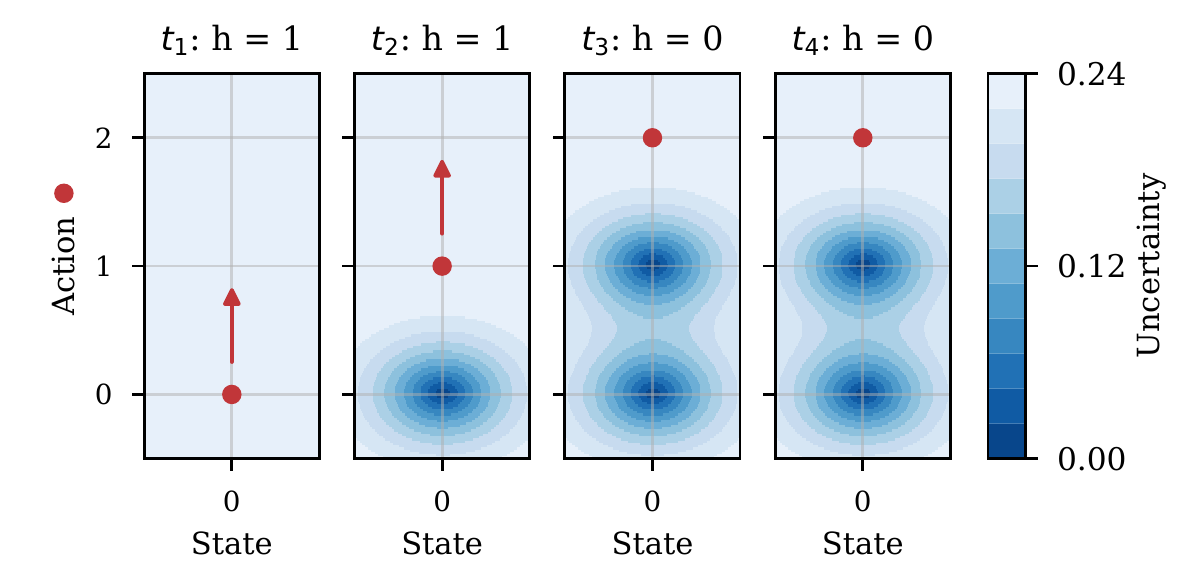}
            \caption{This hypothetical situation clarifies how the action and uncertainty evolves with every feedback sample. The uncertainty is shown for the human model $H$ and is reduced for a $(s,a)$-pair as feedback is obtained.}
            \label{fig:h_uncertain}
        \end{figure}

        \acp{GP} provide uncertainty estimates with every query point based on dissimilarity with the training data. 
        For the policy, the uncertainty reflects the presence of feedback data in the respective or surrounding state.
        Due to the integration of the action in the input of the human model, this uncertainty reflects the presence of feedback for state-action pairs.
        To elaborate on this, a hypothetical example is depicted in \fref{fig:h_uncertain}. 
        The contiguous plots show the evolution of the policy and its uncertainty as new feedback is obtained. 
        We may envision this principle as building
        a map that discloses certain and uncertain regions with respect to past feedback.
        This feature comprises the main advantage of \ac{GPC} over other methods.
        \subsubsection{\acf*{ALR}}\label{sec:learningrate}
    
        We assume that the teacher encounters two teaching phases during the learning period.
        The initial learning phase arises when the process is commenced and the policy is idle. 
        We believe that the feedback in this stage will mostly concern raw adjustments in order to create a coarse version of the final policy.
        These coarse adaptations will gradually shift towards the second learning phase where trainers apply small refinements to the policy for meticulous improvements.
        In this study, we model the transition from coarse to fine adjustments not as a universal annealing process. Instead, we adapt the learning rate to the intended correction per state.

        Hence, we introduce the following \ac{ALR}:
        \begin{equation}\label{eq:learningrate}
            r_k = \sigma_p(\textbf{s}_k) + \sigma_h(\textbf{z}_k)  + c_r,
        \end{equation}
        with $r_k$ the learning rate, $\textbf{s}_k$ the state and $\textbf{z}_k$ the concatenation of $(\textbf{s}_k,a_k)$ (see line 7-9 in \aref{alg:GCOACH}). 
        The uncertainty of the policy $\sigma_p$ allows us to accelerate the learning by increasing the learning rate for the first feedback instances. 
        The uncertainty estimation of $\sigma_h$ adopts a high value for consistent feedback (see \fref{fig:h_uncertain}). As soon as alternating feedback is given, the uncertainty, and thus the learning rate decreases to allow for refinements.
        The parameter $c_r$ denotes the constant rate and prevents stagnation in the event that $\sigma_{p,h} \approx 0$.  
        \ac{GPC} differs from \ac{COACH} for updating the policy, since the error magnitude $e$ is now implicitly included in the computation of $r_k$ in \eqref{eq:learningrate}.
        
\begin{figure}
            \centering
            \includegraphics[trim=7 7 8 7, clip,width=\linewidth]{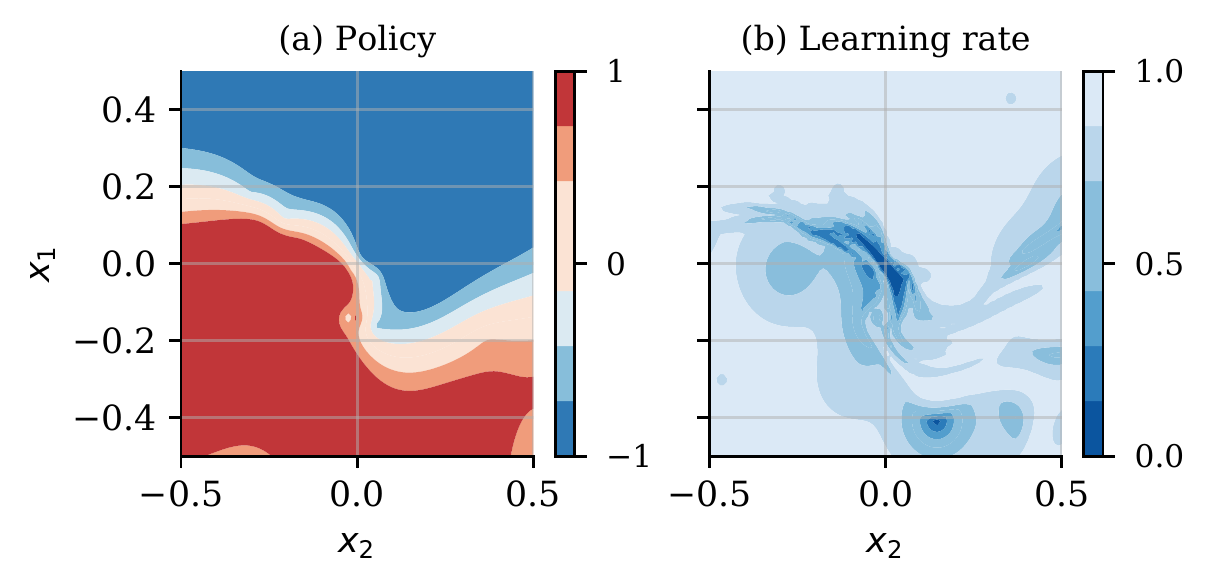}
            \caption{Snapshot of a policy \textit{(a)} and learning rate \textit{(b)} for controlling a system around an unstable equilibrium $(0,0)$.}
            \label{fig:learning_rate}
        \end{figure}

        An example of the policy and the learning rate during a learning process is depicted in \fref{fig:learning_rate}. It shows an environment with a two-dimensional continuous state-space and an unstable equilibrium as reference at $(x_1,x_2)=(0,0)$. 
        The policy \textit{(a)} is trained by a teacher employing the \ac{ALR}. The corresponding learning rate is displayed in \textitt{(b)}. Note that for critical states (area around $(x_1,x_2)=(0,0)$) alternating feedback has caused the \ac{ALR} to decrease, such that the policy can be refined.

        \subsubsection{Active Learning (AL)}\label{sec:activelearning}
        The available uncertainty of the \acp{GP} can be used in an \ac{AL} framework \citep{chernova2014robot}, where high-informative feedback can be inquired for uncertain actions.
        Recent studies have shown great performance improvements with agent-induced feedback, mostly in the \ac{LfD} domain \citep{losey2018including,grave2013learning}. 
        This study is, to the authors' knowledge, the first to assess the potential with a directional feedback framework. 
        Other than \cite{chernova2009interactive}, who employed \ac{AL} with the uncertainty of the visited state, we believe that especially the uncertainty in the action can advance the convergence of directive feedback methods. 
        The motivation for this reasoning is that, in contrast to the \ac{LfD} paradigm, inquiring human assistance in terms of feedback does not yield the optimal action instantaneously. 
        In \ac{GPC} - and feedback implementations in general - multiple feedback instances are needed to approach the optimal policy. Hence, rather than employing state uncertainty, we apply uncertainty per action, which is obtained by the human model 
        \begin{equation}\label{eq:activelearning}
        \Delta_k = c_a\sigma_h(\textbf{z}_k),
        \end{equation}
        with $\textbf{z}_k$ the same as in \eref{eq:humanmodel} and gain $c_a$ to decouple \ac{AL} from the \ac{ALR} (see \eref{eq:learningrate}). 
        By inquiring feedback for high values of $\Delta_k$ we prioritize consistent feedback, since inconsistent feedback would reduce $\Delta_k$. 
        \ac{AL} will therefore further aid in establishing an inaccurate but rather complete policy as early as possible, before proceeding to the refinement stage.
        
        \subsubsection{Sparsification for Corrective Learning}
        \label{sec:policyenhancement}
        
        \begin{algorithm}[t]
        {
          \caption{Sparsification of policy $P$ training data}
          \label{alg:sparsification}
          \begin{algorithmic}[1]
          \Function{Spars}{$\textbf{s}_k$, $a_n$, $\sigma_p$, $\sigma_\text{thres}$, $N_p$}
            \If{$\sigma_p < \sigma_{\text{thres}}$ }
              \State index $\gets \operatorname{arg\,max}_i \: k_p(\textbf{s}_i,\textbf{s}_k)$
              \State $N_p$(index) $\gets (\textbf{s}_k,a_n)$
            \Else 
              \State Append dictionary $N_p =\{\ldots,(\textbf{s}_{k},a_{n})\}$
            \EndIf
            \State \Return $N_p$
            \EndFunction
            \end{algorithmic}
        }
        \end{algorithm}
        
        For every feedback instance provided by the trainer, the dictionary of the policy $P$ is appended with the new tuple:
        \begin{equation}\label{eq:appenddata}
        N_p =\{\ldots,(\textbf{s}_{m+1},a_{m+1})\}
        \end{equation}
        where $(\textbf{s}_{m+1},a_{m+1})$ is calculated based on the executed action $a_k$, learning rate $r_k$ and feedback $h_k$, i.e. 
        \begin{equation}
        a_{m+1} = a_k + r_k h_k.
        \end{equation}
        This approach renders the previous action $a_k$ obsolete. 
        In this application, a deficient property of \acp{GP} that hinders convergence is that by appending the dictionary following \eref{eq:appenddata}, the updated action on $\textbf{s}_{m+1}$ is an average of $a_k$ and $a_{m+1}$ (assuming a coinciding or adjoining data instance). 
        We therefore propose a sparsification method in which the tuple most relevant to the obsolete action $a_k$ is omitted, rendering $a_{m+1}$ the new action. 
        Taking relevancy into account while preserving the uncertainty estimations was not found in conventional online sparsification methods 
        \citep[e.g.][]{nguyen2010incremental}. 
        We therefore introduce a new sparsification technique that specifically applies to applications with iterative updates on the \ac{GP} policy model. 
        
        The main outline of this sparsification is as follows: for every new feedback instance $(\textbf{s}_{m+1},a_{m+1})$, the uncertainty of the policy $\sigma_p(\textbf{s}_k)$ is compared against a certain threshold $\sigma_\text{thres}$.
        We set this threshold to
        \begin{equation}
            \sigma_\text{thres} = \frac{1}{2}\sqrt{\sigma_{s,m}^2},
        \end{equation}
        with $\sigma_{s,m}^2$ either from \eref{eq:sekernel} or \eref{eq:matern}. 
        In the event that this threshold is exceeded,
        the dictionary sample with the biggest covariance (i.e.\ smallest \textitt{Mahalanobis} distance \citep{mahalanobis1936generalized}) is omitted. We thereby prevent the policy from being negatively influenced by obsolete (old) training data.
        %
        %
        The sparsification method is presented in \aref{alg:sparsification} and executed simultaneously with appending $N_p$, see line~11 in \aref{alg:GCOACH}. 
        The existing input elements of the policy dictionary $N_p$ are denoted by $\textbf{s}_i$.
        %
        %
        %

\section{EXPERIMENTAL SETUP}\label{sec:expsetup}
    In this section we detail the experiments in which the performance of \ac{GPC} is evaluated.
    The tests are carried out in three standardized benchmark problems from the OpenAI Gym \citep{openaigym}, namely the Inverted Pendulum(-v0), the Cart-Pole(-v0) and the Lunar Lander(-v2).
    The experiments with oracles (synthesized feedback source) are introduced to test the performance with consistency for all algorithms. 
    The oracle tests also comprise the \ac{AL} and \ac{ALR} assessment. The applicability to the interactive domain is tested in separate experiments with actual human feedback. 
    The performance of \ac{GPC}\footnote{github.com/DWout/GPC} will be tested against baseline \ac{COACH} throughout \citep{celemin2015coach}\footnote{github.com/rperezdattari/COACH-gym}.
            \begin{figure}
                \centering
                \includegraphics[trim=7 50 8 30, clip,width=\linewidth]{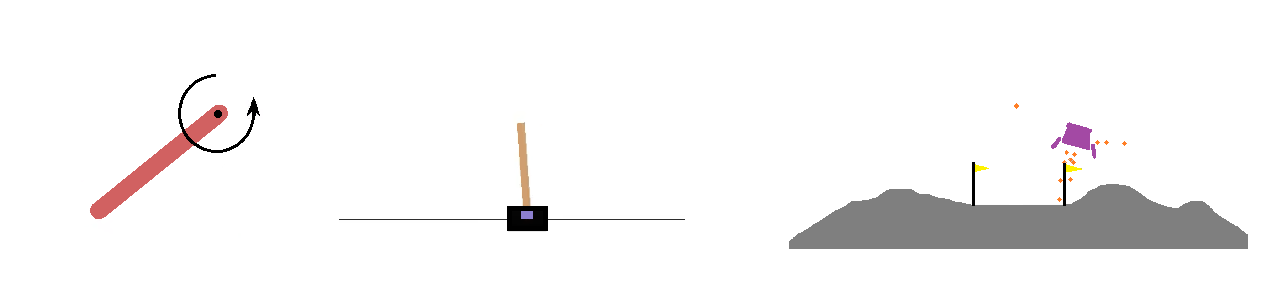}
                \caption{A snapshot of each domain used for the experiments. the most left benchmark denotes the Underactuated Inverted Pendulum(-v0), The Cart-Pole(-v0) environment in the middle, and most right the Lunar Lander(-v2) \citep{openaigym}. The environments are sorted with respect to complexity.}
                \label{fig:envs}
            \end{figure}
        \subsection{ORACLE TESTS}\label{sec:oracle}
        
            An oracle simulates human feedback based on a comparison of the executed action with a reference policy. 
            The oracle experiments are carried out to exclude human factors such as inconsistency and limited attention that hinder a fair comparison between methods. Furthermore, it allows to accurately study the robustness of the algorithms to  erroneous feedback.
        \subsubsection{Performance and Robustness tests}
            First, we will assess the  performance of \ac{GPC}  for perfect and erroneous feedback. 
            For the experiments we set a static feedback rate $\gamma = 5\%$.
            When the action is close to the target action within the range $\delta$, i.e.\ $|a_k - a^*| = \delta$, the policy is considered converged and receives no more feedback.
            The robustness of the algorithms is tested by erroneous feedback. In this study, we will adopt error rates of $10\%$ and $20\%$, which will be administered following the protocol of \cite{celemin2018fast}.
            
        \subsubsection{Active Learning (AL)}\label{sec:activelearningexp}

            The potential regarding \ac{AL} is assessed by encouraging feedback for uncertainty policy actions. 
            In order to exclude any random human factors, the performance is measured using an oracle.  
            As such, we will adapt the feedback rate by incorporating the uncertainty of the human model $H$, i.e.: $\gamma = \Delta_k + \gamma_c$,
            with $\Delta_k$ as in \eref{eq:activelearning} and $\gamma_c$ denoting the minimum feedback rate. 
            The application of \ac{AL} is measured against a baseline with the same episodic feedback rate, but not uncertainty triggered (see \tref{tab:activelearningtests}, \textit{ii} and \textit{iv}). The \ac{ALR} is excluded from the test to circumvent any influences. To account for feedback inconsistencies the erroneous feedback likelihood is set to $10\%$.

        \subsubsection{Ablation Study}
            
            The ablation assessment will analyze the contribution of the \ac{ALR} (\sref{sec:learningrate}). We will run oracle tests employing the learning rate in \eref{eq:learningrate} and compare this to the baseline test from \sref{sec:activelearningexp} with the same episodic feedback rate and erroneous feedback likelihood.
            In addition, we will test a combination of \ac{AL} and \ac{ALR}.
            A summary of the experimental setup is presented in \tref{tab:activelearningtests}.

            
            
            \begin{table}[ht]
            \begin{center}
            \caption{Overview of the experiments regarding Active Learning (AL) and  the ablation study for the Adaptive Learning Rate (ALR). For fair comparison the experiments are conducted with the same feedback (Fb) rate $\gamma$.}
            {\small
            \begin{tabular}{ lcccc }
            \hline \hline
            & AL: & ALR: & Fb rate $\gamma$: & Learning rate $r_k$:\\
            \hline
            \textit{i} & \checkmark & \checkmark& $ \Delta_k + \gamma_c $ & $\sigma_p(\textbf{s}_k) + \sigma_h(\textbf{z}_k)  + c_r$ \\
            \textit{ii} & \checkmark& \xmark & $ \Delta_k + \gamma_c$ & $r_c$ \\
            \textit{iii} & \xmark& \checkmark & $ \gamma_\text{ep.avg}(i)$ & $\sigma_p(\textbf{s}_k) + \sigma_h(\textbf{z}_k)  + c_r$ \\
            \textit{iv} & \xmark&\xmark & $  \gamma_\text{ep.avg}(ii)$ & $r_c$\\
            \hline \hline
    
            \end{tabular}
            }
            \label{tab:activelearningtests}
            \end{center}
            \end{table}

        \subsection{HUMAN TEACHERS}\label{sec:humantests}
            This section will elaborate the experiments for validating the application of \ac{GPC} to interactive settings. 
            The experiments are conducted by employing four human teachers (in the age of 20 to 30, of different background) to the three proposed benchmarks with the objective to achieve convergence as fast as possible. 
            The participants perform the training with every algorithm for every environment four times: two dummy runs to get acquainted with the environment, and two real runs that are recorded. The tests runs are performed single blind: the participants are not informed about which algorithm they controlled. 

\section{EXPERIMENTAL RESULTS}\label{sec:results}
        In this section, we report GPC's performance on the three domains (see \sref{sec:expsetup}). 
        The kernels and \textit{hyperparameters} for the human model and the policy are depicted in \tref{tab:hyper}. 
        For readability purposes the presented results are a walking mean of $3$ samples, unless otherwise specified.

        \begin{table}[ht]
            \begin{center}
            \caption{Hyperparameters of the \acp{GP} in the benchmarks. The policy and human model are modelled by Squared Exponential (SE) and Mat\'ern (Mat.) kernel. The constant learning rate in \eref{eq:learningrate} is denoted as $c_r$.}
            {\small
            \begin{tabular}{ lcccccc } 
            \hline\hline
             & \multicolumn{2}{c}{Pendulum} & \multicolumn{2}{c}{Cart-Pole} & \multicolumn{2}{c}{Lunar Lander} \\
            & CS & NS & CS & NS & CS & NS\\
            \hline
            $H$:                & \textit{SE}          & \textit{SE}      & \textit{SE}    & \textit{SE} & \textit{SE} & \textit{SE}   \\
            \hspace{.1cm} $c_h$      & $0.7$             & $0.45$             & $0.01$         & $0.08$      & $0.01$      & $0.08$                 \\
            \hspace{.1cm} $l_h$      & $0.1$             & $0.1$              & $0.2$          & $0.5$       & $0.2$       & $0.2$                      \\
             $P$:                & \textit{Mat.}        & \textit{Mat.}        & \textit{Mat.}    & \textit{Mat.} & \textit{Mat.} & \textit{Mat.}       \\
            \hspace{.1cm}$c_p$      & $0.01$             & $0.03$             & $0.01$         & $10^{-3}$  & $0.01$      & $10^{-3}$                 \\
            \hspace{.1cm}$l_p$      & $0.7$              & $0.5$              & $0.2$          & $0.7$       & $0.4$       & $0.6$                  \\
            \hspace{.1cm}$\nu_p$    & $0.5$              & $1.5$             & $1.5$         & $1.5$      & $1.5$      & $1.5$                 \\
            $c_r$                     & $0.01$             & $0.02$             & $0.02$         & $0.05$      & $0.02$      & $0.05$             \\
            \hline \hline
            \end{tabular}
            }
            \label{tab:hyper}
            \end{center}
        \end{table}  

        \subsection{PERFORMANCE AND ROBUSTNESS}
\begin{figure}
            \centering
            \includegraphics[trim=7 7 8 7, clip,width=\linewidth]{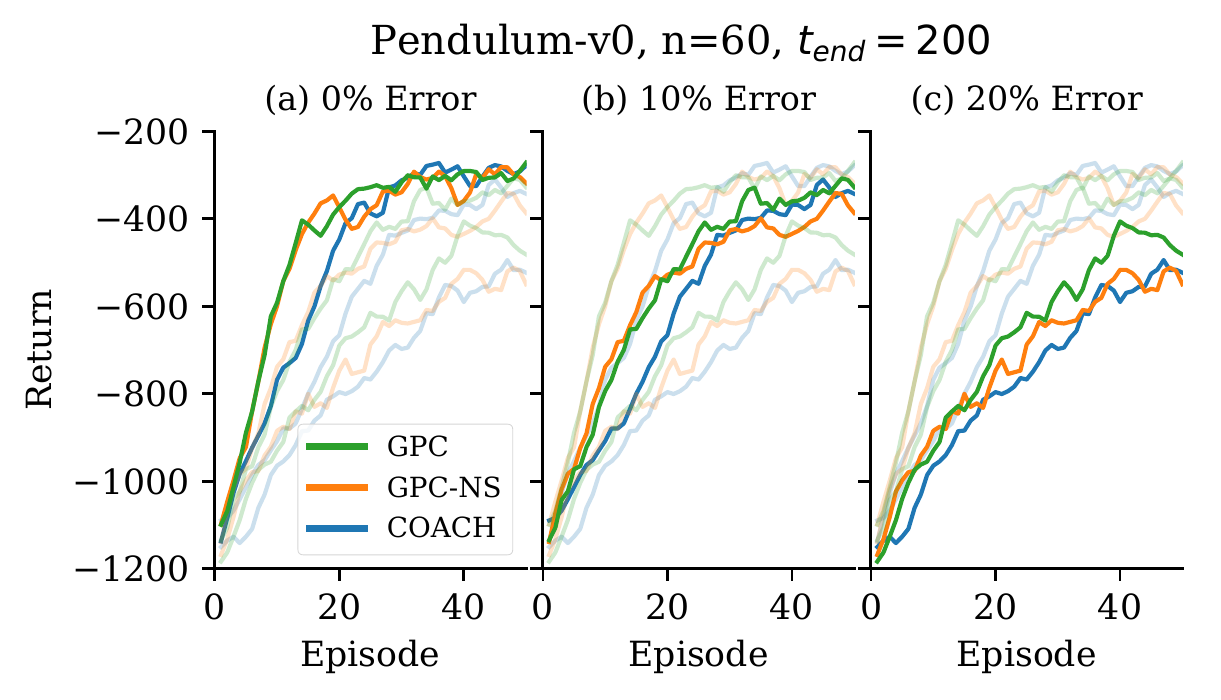}
            \caption{Average environmental return per episode for the Pendulum-v0 domain. \ac{GPC} and \ac{GPC}-NS show nearly the same performance, and outperform \ac{COACH} mainly in the leading domain. The final performance is similar. The performance for every error rate is reprinted (shaded) in all figures for cross-comparison.}
            \label{fig:pendulum_return_2}
        \end{figure}

\begin{figure}
            \centering
            \includegraphics[trim=7 5 8 7, clip,width=\linewidth]{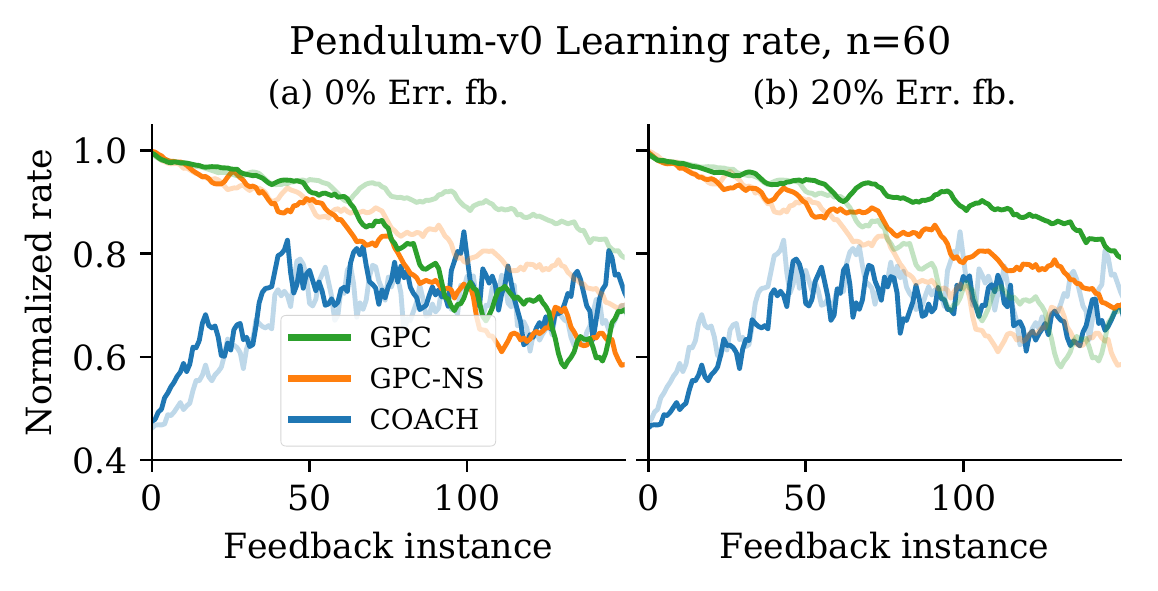}           \caption{Normalized average learning rate in the Pendulum-v0 domain for both \ac{GPC} and \ac{COACH}. In contrast to existing methods, the learning rates of the \ac{GPC} implementations diminish over time, such that the corrections become more subtle upon convergence. As desired, this reduction develops more gradually in case of erroneous feedback.}
            \label{fig:learning_rate_pen}
        \end{figure}

        The return for the Pendulum domain is depicted in \fref{fig:pendulum_return_2}. 
        The \ac{GPC} variants show similar convergence and robustness properties. Due to the coarser exploration in the initial learning phase the learning curve is steeper in comparison to \ac{COACH}. 
        The final performance is similar.
        The average learning rate for all consecutive feedback instances is depicted in \fref{fig:learning_rate_pen} for an error rate of $0\%$ and $20\%$. For \ac{GPC} we see a more aggressive learning rate for the initial learning phase, which diminishes upon convergence. For $20\%$ erroneous feedback this propagation is more gradual, as it should be to reflect on the impeded learning where refinements are appropriate at a later time.

\begin{figure}
            \centering
            \includegraphics[trim=7 7 8 7, clip,width=\linewidth]{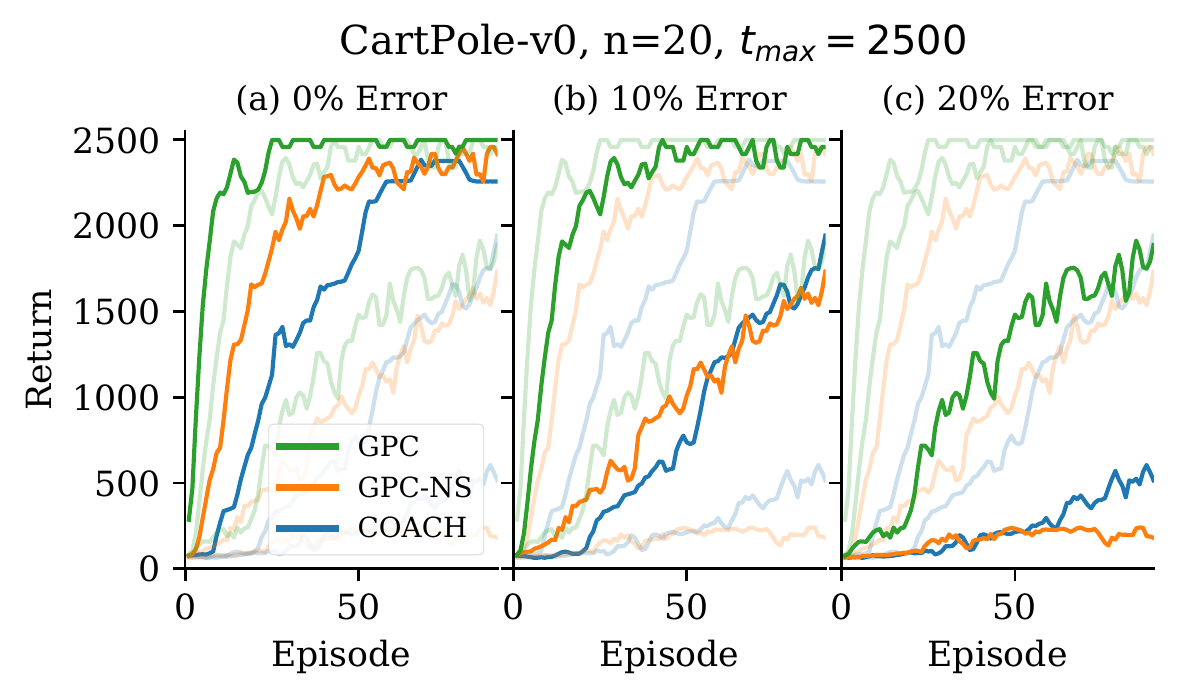}
            \caption{Average return per episode for the CartPole-v0 domain. Both \ac{GPC} implementations outperform \ac{COACH} for ideal feedback. \ac{GPC} shows good robustness to erroneous feedback, whereas \ac{GPC}-NS is more brittle.}
            \label{fig:cartpole_return_2}
        \end{figure}

        The steeper initial learning curve, which was observed in the Pendulum domain in \fref{fig:pendulum_return_2}, also distinguishes \ac{GPC} from \ac{COACH} in the Cart-Pole environment (see \fref{fig:cartpole_return_2}). 
        The protracted take-off time for \ac{COACH} is presumably a result of the \textitt{human feedback supervised learner} module \cite[see][]{celemin2015coach} that adopts a reduced learning rate for the initial learning phase. For every error rate \ac{GPC} outperforms \ac{COACH}. 
        The tuning convenience for \ac{GPC}-NS shows to trade with some sub-optimal performance for erroneous feedback.
        
        
\begin{figure}
            \centering
            \includegraphics[trim=7 7 8 7, clip,width=\linewidth]{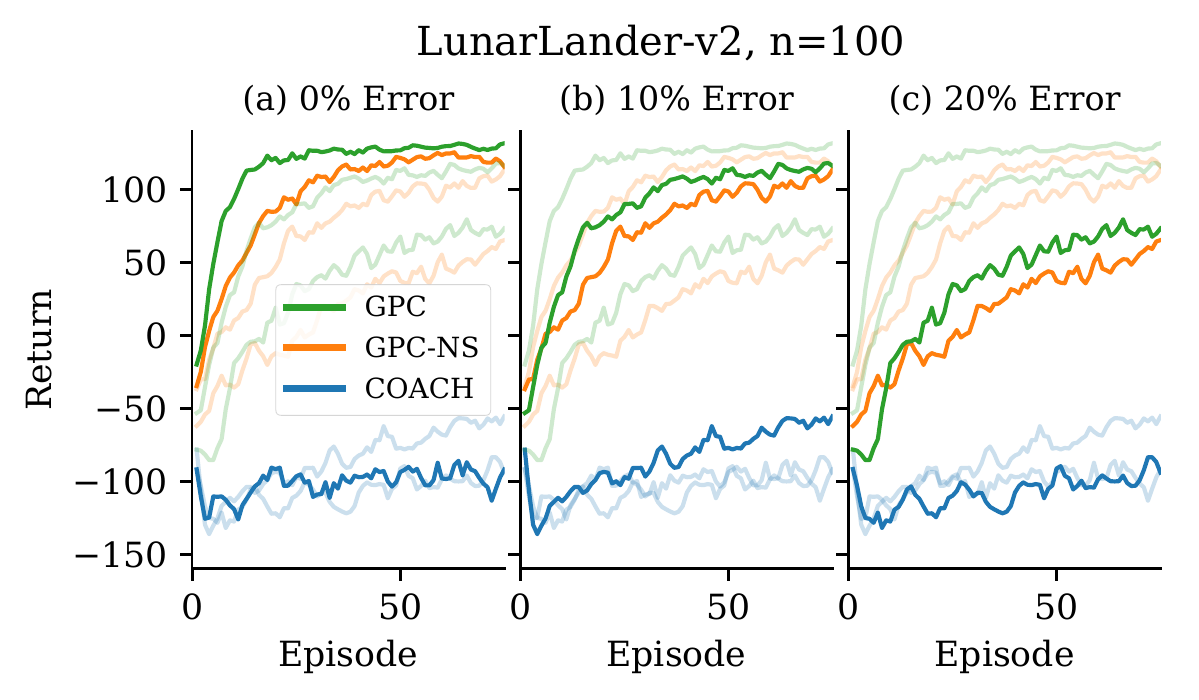}
            \caption{Average return for the LunarLander-v2 environment. Both \ac{GPC} implementations achieve good performance either for ideal or erroneous feedback. \ac{COACH} yields poor performance due to intractability of the feature space, which is custom parameterized in $\mathbb{R}^{24}$.}
            \label{fig:lunar_return_2}
        \end{figure}

        The performance of \ac{GPC} and \ac{COACH} in the Lunar Lander domain is depicted in \fref{fig:lunar_return_2}. Both \ac{GPC} and \ac{GPC}-NS outperform \ac{COACH} for every error rate. \ac{COACH} was unable to achieve any performance due to the unfeasible manual parameterization of the feature space in $\mathbb{R}^{24}$ (lower/upper bound and interval for every input dimension \citep{busoniu2010reinforcement}). 

        \subsection{ACTIVE LEARNING AND ABLATION}
        
\begin{figure}
            \centering
            \includegraphics[trim=7 0 8 7, clip,width=\linewidth]{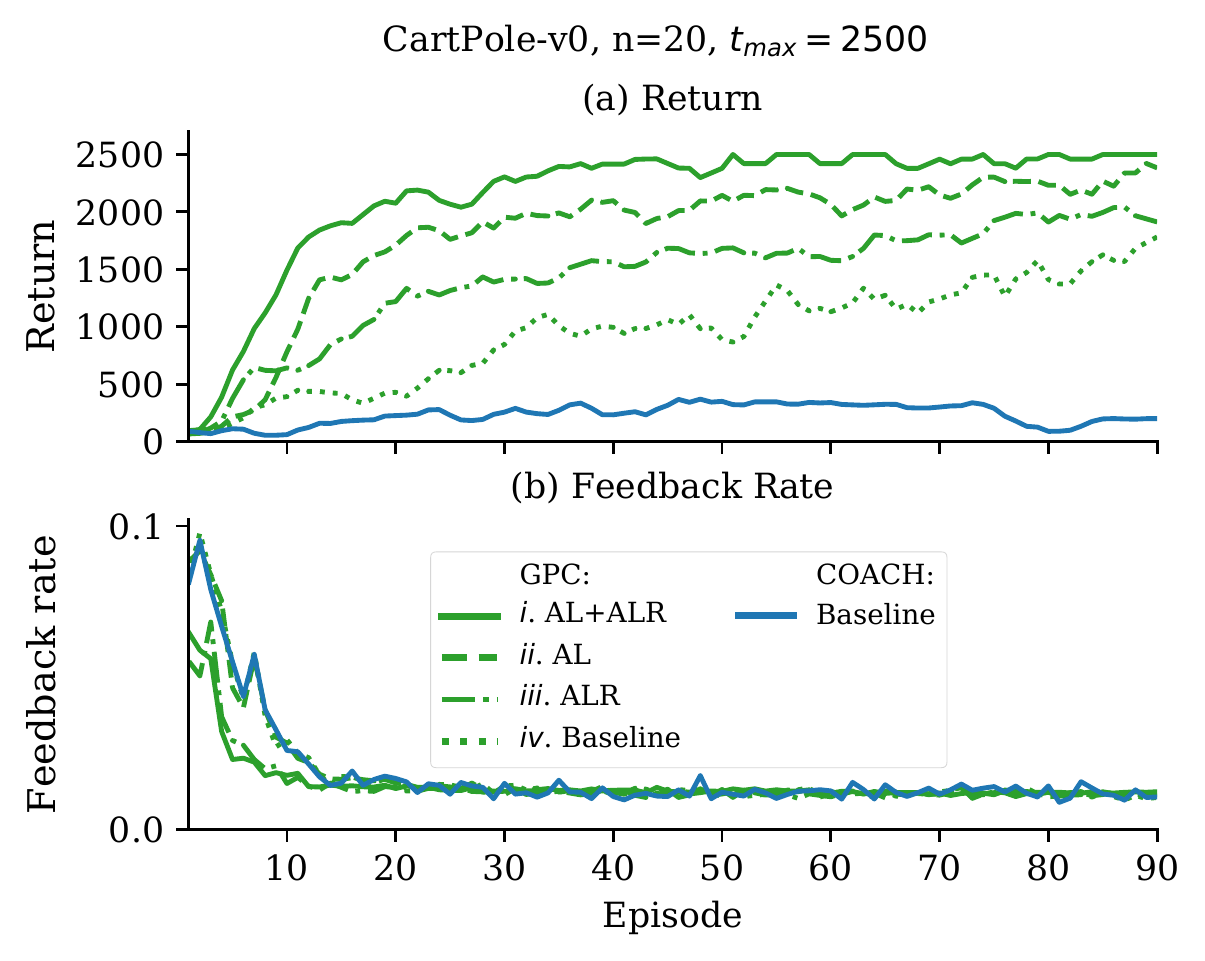}
            \caption{Average return \textit{(a)} and feedback rate \textit{(b)} for the Active Learning (AL) and the ablation of \acf{ALR} in the Cart-Pole domain. AL and ALR combined achieve superior performance. It shows that the Adaptive Learning Rate  accelerates the convergence with less feedback.}
            \label{fig:inquiries}
        \end{figure}

        The result of the four test cases from \tref{tab:activelearningtests} are displayed in \fref{fig:inquiries} for the Cart-Pole environment
        with constant feedback likelihood of $\gamma_c = 0.01$, constant additive learning rate of $c_r = 0.01$ and a static learning rate of $r_c = 0.4$.
        \ac{AL} combined with \ac{ALR} has superior performance. The individual components (\ac{ALR} and \ac{AL} resp.) both prove their significance compared to the baseline. The average learning rate for the \ac{ALR} tests measures $0.0386$ for $i$ and $0.0374$ for $iii$, which is lower on average but better balanced to the static rate of $r_c = 0.4$ in $ii$ and $iv$.
 
        %
    
        \subsection{HUMAN VALIDATION}\label{sec:humanvalidation}
\begin{figure}
            \centering
            \includegraphics[trim=7 5 8 7, clip,width=\linewidth]{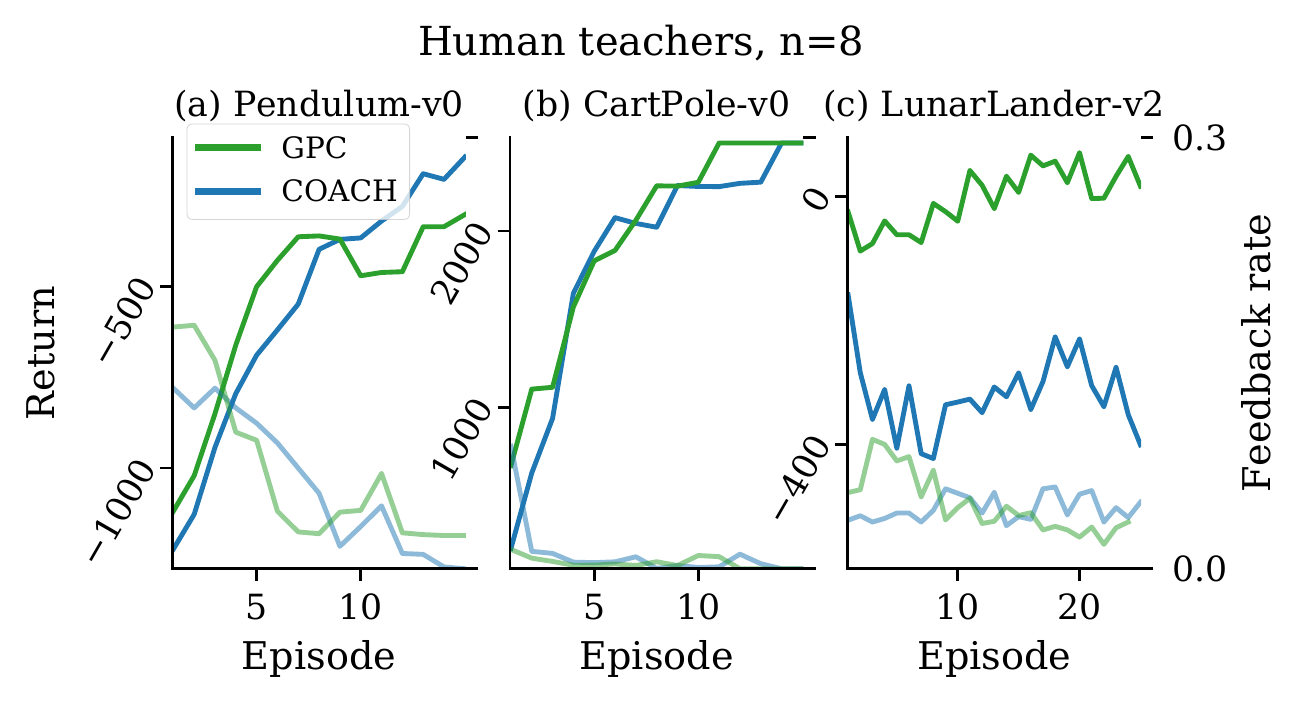}
            \caption{Average return of eight experiments from four human teachers on the three adopted domains.
            Performance is similar to oracle tests and validate the suitability of \ac{GPC} to the interactive domain.}
            \label{fig:human_tests}
        \end{figure}

            The performance for all environments is depicted in \fref{fig:human_tests}. 
            For the Inverted Pendulum and the Cart-Pole environment both \ac{COACH} and \ac{GPC} converge to maximal performance.     
            Although some relative differences are noticeable in the learning curve, 
            the variations are not deemed statistically significant considering the number of testst.            
            The fact that the lacking robustness of the \ac{COACH} implementation (\fref{fig:cartpole_return_2}) does not emerge in this result is notable. 
            In contrast to the static behavior of oracles, 
            humans anticipate to the consequences of the provided feedback and adapt their feedback strategy accordingly. 
            When the corrections at a particular state are deemed insufficient, teachers may choose to provide multiple feedback samples subsequently in order to realize the intended effect.
            An interesting observation is the difference in return for the Lunar Lander environment in \textit{(c)}, 
            which validate the findings from the oracle benchmarks in \fref{fig:lunar_return_2}:
            The unfeasible parameterization of the feature space severely deteriorate the performance in higher dimensional problems. 
\section{CONCLUSION}
\acresetall

Humans are very efficient in understanding control strategies using intuition and common sense.
Corrective feedback is an especially effective means of communication and 
the current state-of-the-art, \ac{COACH},
enables one to establish a control law without requiring control or engineering expertise. 
Moreover, performance is superior over methods that learn autonomously or from evaluative feedback.
However, \ac{COACH} employs \ac{RBF} networks for modelling 
which requires meticulous feature space engineering before these advantages enter into force. 

In this work, we have presented \ac{GPC}. 
It has an architecture similar to \ac{COACH}, but it engages \acp{GP} such that modelling expertise is no longer required and the limitation to confined problems is hereby overcome.
Moreover, we leverage the available uncertainty with an \ac{ALR} that adapts to the trainer's learning phase. In addition, we introduced a new sparsification technique, specifically designed for efficient and accelerated \ac{GP} policy updates.
\ac{GPC} was applied to three continuous benchmarks from the OpenAI Gym: the Inverted Pendulum, Cart Pole, and Lunar Lander. 
Our novel framework outperform \ac{COACH} on every domain tested by means of faster learning and better robustness to erroneous feedback. 
The greatest improvement was for the Lunar Lander problem, where RBF parameterization fails but \ac{GPC} is flawless.

In addition to the performance and robustness assessment, we performed two additional studies:
1) 
We have addressed the potential of \ac{AL} and demonstrated how eliciting feedback for actions with greatest uncertainty yields drastic improvements on convergence.
2)
We have furthermore presented an alternative implementation 
\ac{GPC}-NS where length-scale tuning is circumvented by online normalization of the input space. 
It is slightly suboptimal and trades some robustness in comparison to \ac{GPC}, but a great advantage is that it does not require any input parameterization in new domains.
This could especially be decisive in higher-dimensional problems, and furthermore renders our work feasible also for non-experts.

In future work, it might be possible to further innovate the dynamical scaling, such that the applicability and generality of \ac{GPC}-NS is again extended.
In addition, the \ac{AL} opportunities assessed here deserve further research and should be validated also with human participants. 


\begin{acronym}[TDMA]
	\acro{A-OPI}{Advice-Operator Policy Improvement}
    \acro{AL}{Active Learning}
    \acro{ALR}{Adaptive Learning Rate}
    \acro{ARD}{Automatic Relevance Determination}
    \acro{CEM}{Cross-Entropy Methods}
    \acro{CMA-ES}{Covariance Matrix Adaption - Evolution Strategies}
    \acro{CMA}{Covariance Matrix Adaption}
	\acro{COACH}{COrrective Advice Communicated by Humans}
    \acro{CS}[CS]{Custom Scaling}
    \acro{GP}{Gaussian Process}
	\acro{GPC}{Gaussian Process Coach}
    \acro{LfD}{Learning from Demonstration}
	\acro{LFD}[LfD]{Learning from Demonstration}
    \acro{ML}{Machine Learning}
    \acro{MDP}{Markov Decision Process}
    \acro{NS}[NS]{Normalized Scaling}
    \acro{PG}{Policy Gradient}
    \acro{PS}{Policy Search}
    \acro{RL}{Reinforcement Learning}
        \acrodefplural{GP}{Gaussian Processes}
        \acrodefplural{MDP}{Markov Decision Processes}
    \acro{RBF}{Radial Basis Function}
        \acrodefplural{RBF}{Radial Basis Functions}
    \acro{SABL}{Strategy-Aware Bayesian Learning}
	\acro{TAMER}{Training an Agent Manually via Evaluative Reinforcement}
\end{acronym}

\bibliographystyle{apalike}
\bibliography{paper}



\end{document}